\pdfoutput=1
\documentclass{article}

\usepackage{amsmath}
\usepackage{spconf,graphicx}

\usepackage{flushend}
\usepackage[colorinlistoftodos]{todonotes}
\usepackage{hyperref}
\hypersetup{colorlinks,allcolors=blue}
\usepackage{cite}
\usepackage{booktabs} 
\usepackage{balance}
\usepackage{fancyhdr}
\usepackage{courier}  
\usepackage{url}  
\usepackage[ruled,lined,vlined,linesnumbered]{algorithm2e}
\usepackage[noend]{algorithmic}
\usepackage{bm}
\usepackage{url}
\usepackage{enumitem}
\usepackage{changepage}
\usepackage{color}
\usepackage{multirow}
\usepackage{amsfonts} 
\usepackage{amssymb}
\usepackage{float}
\usepackage{soul}
\usepackage{subfig}
\usepackage{cleveref}
\usepackage{bbm}
\usepackage{footnote}
\usepackage{amssymb}
\makesavenoteenv{tabular}
\usepackage{threeparttable}
\usepackage{pgfplots}

\newcommand{\rr}[1]{{\color{blue}[Ruirui: #1]}}
\newcommand{\YC}[1]{{\color{red}[Yuguang: #1]}}
\newcommand{\ZC}[1]{{\color{orange}[Zeya: #1]}}

\title{Self-supervised Speaker Recognition Training Using \mbox{Human-Machine Dialogues}}
\name{Metehan Cekic$^{\S}$\sthanks{Work done during an internship at Amazon.}
\quad Ruirui Li$^{\dagger}$ \quad Zeya Chen$^{\dagger}$ \quad Yuguang Yang$^{\dagger}$
\quad Andreas Stolcke$^{\dagger}$ \quad Upamanyu Madhow$^{\S}$}

\address{$^{\S}$University of California, Santa Barbara, CA, U.S.A. \quad $^{\dagger}$Amazon Alexa AI, Sunnyvale, CA, U.S.A.\\
			\{metehancekic, madhow\}@ece.ucsb.edu \quad
			\{ruirul, zeyachen, yuguay, stolcke\}@amazon.com} 

\pgfplotsset{compat=1.17}

\begin{document}
\ninept
\maketitle
\begin{abstract}
Speaker recognition, recognizing speaker identities based on voice alone, enables important downstream applications, such as personalization and authentication.
Learning speaker representations, in the context of supervised learning, heavily depends on both clean and sufficient labeled data, which is always difficult to acquire.
Noisy unlabeled data, on the other hand, also provides valuable information that can be exploited using self-supervised training methods. 
In this work, we investigate how to pretrain speaker recognition models by leveraging dialogues between customers and smart-speaker devices.
However, the supervisory information in such dialogues is inherently noisy, as multiple speakers may speak to a device in the course of the same dialogue.
To address this issue, we propose an effective rejection mechanism that selectively learns from dialogues based on their acoustic homogeneity. 
Both reconstruction-based and contrastive-learning-based self-supervised methods are compared.
Experiments demonstrate that the proposed method provides significant performance improvements, superior to earlier work. Dialogue pretraining when combined with the rejection mechanism yields $27.10$\% equal error rate (EER) reduction in speaker recognition, compared to a model without self-supervised pretraining. 
\end{abstract}

\begin{keywords}
self-supervised training, speaker recognition, dialogue, rejection mechanism
\end{keywords}

\section{Introduction}
\label{sec:intro}

\begin{table*}[t]
\centering
\caption{Sample dialogues}
\scalebox{0.9} {
\begin{tabular}{ c | l | l | l | l }
\toprule
Dialogue ID & Device type & Time & Source & Utterance\\ \hline \hline
\multirow{4}{*}{A} & \multirow{4}{*}{Google Home} & 2021-07-03 12:12:02 & Customer & Hey Google, what's the weather like tomorrow? \\ \cline{3-5}
& & 2021-07-03 12:12:12 & Device & In New York city, it will be mostly sunny with the highest 77 and lowest 64.\\ \cline{3-5}

 & & 2021-07-03 12:12:20 & 	Customer & Thanks Google.\\ \cline{3-5}
& & 2021-07-03 12:12:27 & Device & No problem.\\ \hline \hline

\multirow{4}{*}{B} & \multirow{4}{*}{Echo} & 2020-07-01 09:10:01 & Customer & Alexa, add eggs to my shopping list. \\ \cline{3-5}
& & 2020-07-01 09:10:08 & Device & A dozen of eggs of organic large brown eggs have been added into your cart.\\ \cline{3-5}

 & & 2020-07-01 09:10:15 & Customer & Alexa, I also want dark chocolate. Can I have that, Daddy?\\ \cline{3-5}
& & 2020-07-01 09:10:25 & Device & Sorry, I did not recognize your voice. Would you like to get enrolled?\\ 
\midrule
\end{tabular}
}
\label{table:sample_dialogue}
\end{table*}

\begin{figure*}[t]
    \centering
    \includegraphics[width=0.99\textwidth]{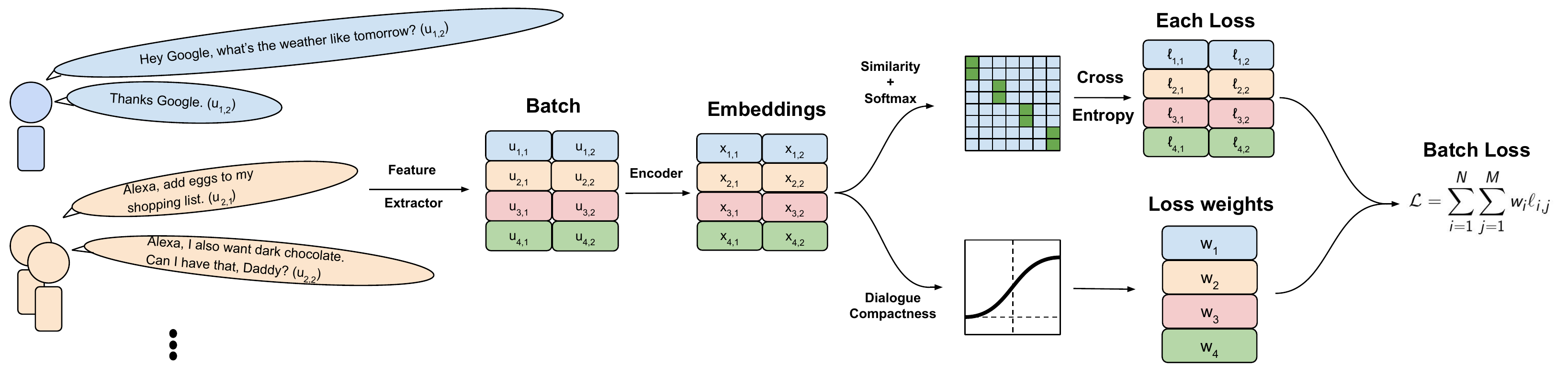}
    \caption{Each batch contains $M=2$ utterances from $N=4$ different dialogues, $M\cdot N$ utterances in total. We multiply the loss from each utterance depending on the compactness of the dialogue the utterance is extracted from. }
    \label{fig:framework}
\end{figure*}

Speaker recognition answers the fundamental question ``who is speaking'' based on a sample of speech, thereby enabling both personalization and authentication in speech-based applications. Most speaker recognition model training is supervised, in the sense that it relies on both clean and sufficient speaker-labeled data~\cite{chung2020defence,chung2018voxceleb2,xie2019utterance,DBLP:conf/wsdm/LiJLH020,DBLP:conf/icassp/LiJWMH020,DBLP:conf/interspeech/LiJWHS20}. However, labeling data  in the quantities required for production-level models is a substantial bottleneck.
Recent work~\cite{DBLP:conf/icassp/SaeedGZ21, DBLP:journals/corr/abs-2012-06659,DBLP:conf/icassp/ChungG20,DBLP:conf/icassp/LiuYCHL20,DBLP:journals/taslp/LiuLL21,DBLP:journals/corr/abs-2106-07447,DBLP:journals/corr/abs-1807-03748} is turning to unlabeled data for pretraining a speech model first, and then fine-tuning it on a smaller labeled dataset.
In this work, we focus on how to pretrain a speech model suitable for speaker recognition tasks.

In general, there are two types of self-supervised methods to pretrain speech models, namely, based on reconstruction or based on contrastive learning.
For the former, one or a few consecutive frames are masked and then models are trained to reconstruct or predict the original features, such as APC~\cite{DBLP:conf/icassp/ChungG20}, MockingJAY~\cite{DBLP:conf/icassp/LiuYCHL20}, DeCoAR~\cite{ling2020decoar} and HuBERT~\cite{DBLP:journals/corr/abs-2106-07447}.
As the masked features are reconstructed based on context, reconstruction-based methods are more suitable to speech recognition tasks and less effective on speaker recognition tasks.
For the latter, positive and negative instances are constructed and models are optimized by conducting comparisons, which aim to group positive instances together while separating negative instances, such as COLA~\cite{DBLP:conf/icassp/SaeedGZ21}, CPC~\cite{DBLP:journals/corr/abs-1807-03748}, and wav2vec~\cite{DBLP:conf/interspeech/SchneiderBCA19}.
Therefore, to effectively distinguish utterances from different speakers, contrastive learning methods are more appropriate.

In this work, we propose a contrastive self-supervised method specialized for speaker recognition. 
To achieve this goal, we leverage unlabeled dialogues between smart-speaker devices and their users.
Table~\ref{table:sample_dialogue} shows two dialogue samples, where each dialogue is composed of spoken interactions between customers and a smart speaker. 
We presume that most dialogues, such as Dialogue A, are clean in the sense that each involves customer utterances from a single speaker only. 
Therefore, customer utterances from the same dialogue serve as positive instances, while customer utterances from different dialogues form negative instances.
However, a few dialogues, e.g., Dialogue B, are noisy with respect to speaker identities, i.e., contain customer utterances from more than one speaker.
In order to avoid contaminating the model training with positive samples involving different speakers, we develop a rejection module. 
The rejection module allows the model to effectively learn from clean dialogues and give less weight to the noisy ones, leading to a more accurate and robust speaker recognition model.

Thus, our contribution is a self-supervised learning method for speaker recognition systems, demonstrating that
\begin{itemize} [label={$\bullet$}, topsep=0pt, itemsep=0pt]
\item a dialogue dataset from human-device interactions is an effective unlabeled data source that can be leveraged in self-supervised pretraining of speaker recognition models;


\item self-supervised rejection is a very effective tool to deal with false positive pairs caused by multi-speaker dialogues, providing more than 15\% equal error rate (EER) improvement even without fine-tuning;


\item fine-tuning the pretrained model utilizing our framework can further improve speaker recognition and relative EER improvement is as high as 41.28\%.

\end{itemize}
\section{Method}
\label{sec:methodology}

The unlabeled dialogue data is noisy because the customer utterances in the same dialogue can come from different speakers. It follows that the positive instances constructed by pairing utterances from the same dialogue are not reliable all the time. To alleviate this issue, we propose a new all-versus-all loss function and a rejection mechanism. 
Unlike angular prototypical loss \cite{chung2020defence} and GE2E loss \cite{Wan2018}, all-versus-all loss avoids using a centroid to represent a dialogue, but rather conducts comparisons for each utterance in a dialogue. In this way, the model will suffer less from centroids aggregating multiple speakers while learning from all utterances in a dialogue.
In addition, the rejection mechanism guides the model in learning more from clean dialogues instead of noisy ones by weighting their loss contributions differently.

Figure~\ref{fig:framework} shows the proposed modeling framework. Given $N$ dialogues ($N=4$ in our case), we randomly sample $M$ customer utterances per dialogue ($M=2$ here) to construct a batch of $M \cdot N$ utterances. An encoder is employed to extract an embedding for each utterance in the batch. Then, a loss for each utterance is calculated based on its similarities to other utterances in the same dialogue and those in other dialogues, which are stored in a similarity matrix. At the same time, a compactness score is calculated for each dialogue, expressing the speaker purity of the dialogues. The overall batch loss is defined by a weighted sum of utterance losses  considering the dialogue compactness scores. 

\subsection{All-versus-All Loss} 
\label{subsec:loss}

\begin{figure*}[tb]
    \centering
    \includegraphics[width=0.9\textwidth]{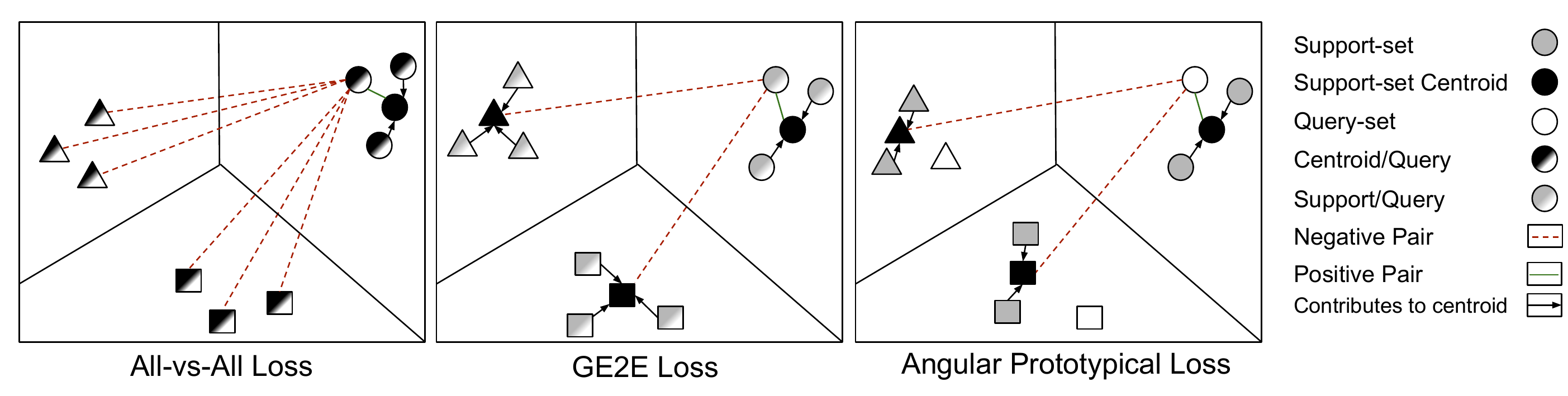}
    \caption{Comparison between the effect of all-versus-all loss (AvA), generalized end-to-end loss (GE2E), and angular prototypical loss (AP). Dashed lines represent distances encouraged to increase, while solid lines represent distances being decreased. Centroids denoted by black nodes are computed as the mean of the support set during training.}
    \label{fig:loss_funcs}
    \vspace{-0.3cm}
\end{figure*}

The presence of the multiple speaker dialogues in the dataset causes the class centroids of GE2E loss to be flawed, as different speakers will have completely different embeddings. Specifically, if there are multiple speakers in a dialogue, the negative pair centroids would not be ideal as the centroids are not reliable and the aggregation step leads to information loss, causing wrong gradient directions. In this work, we propose all-versus-all (AvA) loss function to alleviate the negative pair centroid problem. We compare the embedding to all the other embeddings without relying on centroids, not only avoiding the flawed centroid problem but also increasing the effective number of negative pairs. 
For positive pairs, we compare the query embedding with the centroid of the embeddings coming from the same dialogue, excluding the query utterance. The similarity between the utterances coming from same dialogue is promoted whereas the similarity between the utterances coming from different dialogues is penalized. 
If we form our batch with $M$ utterances coming from $N$ different dialogues, then the number of comparison pairs will be $M \cdot (N-1) + 1$ with only one of them being positive. GE2E and angular prototypical loss, on the other hand, have $N-1$ negative pairs and a single positive pair. \cite{chen2020simple} demonstrated that larger batch sizes, meaning more negative pairs, help improve the performance of self-supervised learning; therefore AvA gives a better loss function for our problem. Figure \ref{fig:loss_funcs} visualizes all the loss functions considered and explains the main differences among them.
Formally,
\begin{equation}
\label{equ:encoder}
\begin{aligned}
x_{i,j}&=f(u_{i,j})
\end{aligned}
\end{equation}
\begin{equation}
\label{equ:cosine}
\begin{aligned}
s(x_{i,j},x_{k,l})&=\frac{x_{i,j}^Tx_{k,l}}{||x_{i,j}||_2||x_{k,l}||_2}
\end{aligned}
\end{equation}
\begin{equation}
\label{equ:compactness}
\begin{aligned}
c_{i,j} &= \frac{1}{M-1}\sum_{k=1,k\neq j}^M x_{i,k}
\end{aligned}
\end{equation}
\begin{equation}
\label{equ:loss_func}
\begin{aligned}
\ell_{i,j} &= -\log \frac{e^{s(x_{i,j},c_{i,j})}}{\sum_{k,l,k\neq i} e^{s(x_{i,j},x_{k,l})}+ e^{s(x_{i,j},c_{i,j})}}
\end{aligned}
\end{equation}
where $f$ is an encoder model that produces embedding representation $x_{i,j}$ from an utterance $u_{i,j}$, where $u_{i,j}$ is the $j$-th utterance from dialogue $i$. $s$ calculates the cosine similarity between two embeddings. $c_{i,j}$ is the positive centroid of a given utterance $u_{i,j}$ based on dialogue $i$. $\ell_{i,j}$ is the cross entropy loss for a given utterance $u_{i,j}$. 

\subsection{Self-supervised Rejection}
\label{subsec:rejection}

Although AvA loss is solving part of the negative pair centroid problem, it does not offer a solution for the positive pair errors caused by the multiple speaker dialogues. Since the model will try to distinguish all the provided training data as well as possible, multi-speaker examples may push the neural network to learn some non-robust and SID-unrelated features. This is similar to the problem of noisy labels in supervised learning \cite{sukhbaatar2014learning, wang2017multiclass}. We employ the idea of loss reweighting to decrease the contributions from noisy dialogues. Our method works on the fly, providing single-pass training with a significant performance improvement. 
We believe this framework also works for speaker recognition with datasets with noisy labels.
Furthermore, our rejection mechanism works without significant additional computational cost since the similarities are already computed for the loss functions.

Given the dialogue embeddings, we compute the average of the pairwise cosine similarities, which we call the {\em compactness} of the given dialogue. Then, we pass the compactness values through a sigmoid function with two hyper-parameters: temperature and threshold. 

\begin{equation}
\begin{aligned}
C_i &= \frac{1}{M(M-1)}\sum_{j=1}^M\sum_{k=1,k\neq j}^M s(x_{i,j}, x_{i,k})
\end{aligned}
\end{equation}
\begin{equation}
\begin{aligned}
w_i &=  \sigma(T*(C_i-t)) 
\end{aligned}
\end{equation}
where $T$ is the temperature controlling the steepness of the sigmoid function, and $t$ is a predefined threshold that determines the center of the sigmoid $\sigma(\cdot)$. We allow $T$ to be learned by the model; however, we do not propagate derivatives through compactness $C_i$, letting it function simply as a scaling factor for the loss values.

Our final loss function becomes the weighted sum of the per-utterance losses:
\begin{equation}
\mathcal{L} = \sum_{i,j} w_{i} \cdot \ell_{i,j} 
\end{equation}
When the similarity between two utterances coming from the same dialogue is relatively small the weight of that particular dialogue will also be small.
The idea is decreasing the loss contribution from multi-speaker dialogues, since their compactness will be much lower than that of single-speaker dialogues.

\section{Experiments}
\label{sec:experiments}

Across all our experiments we employ a model consisting of a multi-layer unidirectional LSTM followed by a fully connected layer network. The dimensionality of each LSTM layer is 768, whereas the last linear layer has 256 units. 

We pretrain our models on the AWS platform using 8 NVIDIA V100 GPUs with 16GB memory for 200 epochs. 
We employ the Adam optimizer with an initial learning rate of 0.0004, decreasing by 2\% every 10000 iterations.
For all of the experiments, we save the model giving the best validation EER value on a small subset of the labeled dataset.

\begin{table*}[t]
\caption{Pretraining and fine tuning results, our method outperforms all the other pretraining methods significantly over the speaker recognition task.}
\vspace{-0.2cm}
\centering
\subfloat[Pretraining results. For each loss function, improvements relative to batch size 32 without rejection are shown.]{
\scalebox{0.82} {
\begin{tabular}{@{}lcccc@{}}
\toprule
\multirow{2}{*}{Loss}  & \multicolumn{4}{c}{Batch Size}          \\ \cmidrule(l){2-5} 
                       & 32      & 64      & 128      & 256      \\ \midrule
All-vs-All             & 0.00\%  & +2.91\% & +6.56\%  & \textbf{+8.20\%}   \\ 
Rejection + All-vs-All & \textbf{+3.76\%} & +7.65\% & +18.76\% & \textbf{+19.00\%} \\ 
A-Proto                & 0.00\%  & +7.32\% & +8.52\%  & +12.93\%   \\ 
Rejection + A-proto    & +7.55\% & +12.58\% & +16.76\% & +25.85\% \\ 
GE2E                   & 0.00\%  & +3.24\% & +3.06\%  & +6.36\%   \\ 
Rejection + GE2E       & +10.64\% & +17.75\% & +17.99\% & +13.83\% \\ \bottomrule
\end{tabular}
\label{table:pretraining}}
}
\hspace{0.5cm}
\subfloat[Fine-tuning results. For all experiments we take the model trained from scratch as our baseline and report the relative improvement.]{
\scalebox{0.72} {
\begin{tabular}{@{}lcccccc@{}}
\toprule
\multirow{2}{*}{Pretraining} & \multirow{2}{*}{Loss} & \multirow{2}{*}{Episodes} & \multicolumn{4}{c}{Labeled Dataset Speaker Count} \\ \cmidrule(l){4-7} 
                              &                       &                           & 1,024         & 2,048       & 4,096      & 8,192      \\ \midrule
-                             & GE2E                  & 1000                      & 0.00\%       & 0.00\%     & 0.00\%    & 0.00\%    \\ \midrule
COLA                          & GE2E                  & 300                       & -8.81\%      &   -23.57\%   &  -37.07\%   & -44.21\%   \\ 
APC                           & GE2E                  & 300                       & +24.34\%     &   +23.13\%   &   +19.48\%   &  +15.35\%    \\ 
VoxCeleb2                     & GE2E                  & 300                       & +31.38\%     &   +25.91\%  & +20.95\%   &   +15.61\%  \\ 
Dialogue+AvA (ours)           & GE2E                  & 300                       & +40.18\%     &    +34.19\%   &   \textbf{+31.10\%} &    \textbf{+27.10\%}        \\ 
Dialogue+A-Proto (ours)       & GE2E                  & 300                       &  \textbf{+41.28\%}     &     \textbf{+34.77\%}  &   +30.03\%  &   +26.57\%        \\ 
Dialogue+GE2E (ours)          & GE2E                  & 300                       & +40.12\%     &    +32.86\%   &  +27.49\%    &   +23.42\%        \\ \bottomrule
\end{tabular}
\label{table:finetuning}
}
}
\vspace{-0.5cm}
\end{table*}

The pretraining is conducted on deidentifed speech dialogues.
The dataset is composed of 927,000 dialogues, comprising about 1800 hour of speech data.
Since the number of dialogues is large, the chance of having multiple dialogues from the same speaker is very low per batch. As a dialogue contains at least two customer utterances, we form each batch by collecting two utterances from $N$ different dialogues. We conduct our experiments using three different loss functions: GE2E, all-versus-all, and angular prototypical. Moreover, in order to investigate the effect of the rejection mechanism we conduct a number of experiments with varying batch sizes. 

The evaluation dataset is constructed by first randomly sampling de-identified utterances from a year's traffic. 
Then each sampled utterance and the enrollment data of speakers are sent to multiple annotators to obtain ground-truth labels independently.
To reduce annotation errors, we select utterances that have consistent annotation labels for the final evaluation dataset.

\subsection{Model Performance with Rejection Mechanism}
We first investigate how the rejection mechanism helps us learn from the noisy unlabeled dialogue data.
Table~\ref{table:pretraining} reports relative EER improvements by taking a batch size of 32, without using rejection, as a baseline. 

There are two observations. First, the rejection mechanism helps improve EER performance on all three loss functions and different batch sizes. For example, when all-versus-all loss is applied and the batch size is 32, we observe 3.76\% relative EER improvement.
This demonstrates the effectiveness of the rejection mechanism, helping the model focus on clean dialogues rather than noisy ones.
Second, a large batch size also contributes to better EER performance, especially when the rejection mechanism is applied. For example, when all-versus-all loss is applied, the EER improves by 8.2\% by increasing the batch size from 32 to 256. It improves by 19.0\% if the rejection mechanism is involved. A large batch involves utterances from more dialogues and forces the model to learn harder tasks, distinguishing more speakers in a batch. This results in more accurate speaker recognition.

\subsection{Model Performance before Fine-tuning}
\begin{table}[tb]
\caption{Comparison of pretrained models, our method outperforms reference model trained on the VoxCeleb2 labeled dataset without fine-tuning. Its performance is even comparable to fully supervised models trained on labeled Alexa datasets.}
\centering
\scalebox{1.0} {
\begin{tabular}{@{}lccr@{}}
\toprule
Training Data & Method type      & Loss    & EER \\ \midrule
Alexa Dialogue      & Self-supervised & COLA    & -129.56\% \\ 
Alexa Dialogue      & Self-supervised & APC     & -108.32\% \\ 
VoxCeleb2     & Supervised   & GE2E    & 0\% \\ 
Alexa (1024 spk)      & Supervised & GE2E     & +12.75\% \\ 
Alexa (2048 spk)      & Supervised & GE2E     & +27.11\% \\ 
Alexa (4096 spk)      & Supervised & GE2E     & \textbf{+34.79}\% \\ 
Alexa (8192 spk)      & Supervised & GE2E     & +39.17\% \\ 
Alexa Dialogue      & Self-supervised & AvA     & +28.81\% \\ 
Alexa Dialogue      & Self-supervised & A-Proto & \textbf{+30.84}\% \\ 
Alexa Dialogue      & Self-supervised & GE2E    & +28.49\% \\ \bottomrule
\end{tabular}
}
\vspace{-0.5cm}
\label{table:precompare}
\end{table}

In this section, we investigate the performance of speaker recognition without fine-tuning the pretrained models.
To compare with other self-supervised methods, we also pretrain a reconstruction-based APC model~\cite{DBLP:conf/icassp/ChungG20}, and COLA~\cite{saeed2021contrastive} based on contrastive learning. As there are several public labeled datasets, we also train a supervised model based on the VoxCeleb2 dataset~\cite{chung2018voxceleb2} to serve as an additional pretrained model. Here the supervised pretrained model based on the VoxCeleb2 dataset serves as the reference.
In addition, we further train four fully supervised models based on labeled Alexa datasets with varying number of speakers.

We highlight three observations based on Table~\ref{table:precompare}. First, we note that the pretrained models COLA and APC are worse than the supervised model trained on the VoxCeleb2 dataset. These two methods aim to learn general audio features and they strongly depend on fine-tuning steps in order to achieve comparable performance for a downstream task. Therefore, they perform poorly on speaker recognition task without fine-tuning.
Second, the proposed model and its variants consistently outperform the reference model trained on the VoxCeleb2 labeled dataset, with EER reduced by as much as 30.84\% relative. This clearly demonstrates the effectiveness of the proposed model in exploiting implicit speaker information in human-machine dialogues.
The utilization of Alexa human-machine dialogues helps us overcome the domain mismatch between Alexa users and speech from other sources, such as the YouTube excerpts assembled in VoxCeleb.
Third, the proposed model achieves EER reductions comparable to the models trained from scratch on Alexa labeled datasets.
For example, our best performing model achieves 30.84\% EER reduction while the fully supervised model trained on the Alexa labeled 4096-speaker dataset achieves 34.79\% reduction. 
This shows that the proposed model trained with unlabeled dialogue data is effective in learning speaker identity features. 

\subsection{Model Performance after Fine-tuning}

We fine-tune the pretrained network on different labeled Alexa datasets with varying number of speakers, where the total utterance duration for a speaker is around 150 seconds on average.
All fine-tuning results based on the various pretrained models are summarized in Table \ref{table:finetuning}. Here the four models trained from scratch with 1024, 2048, 4096, and 8192 labeled speakers serve as the reference baselines.
Due to limited space, we show the fine-tuned model performance for GE2E loss only.


There are four key observations.
First, the pretrained COLA model is not effective at learning speaker identities on the dialogue data, as we observe performance drop compared to the model trained from scratch for all four fine-tuning datasets.
The utterances in dialogues are very short (one to two seconds duration). COLA further separates each utterance into two segments in order to form positive instances. Moreover, the background environment tends to be identical within the same utterance. Without massive and effective data augmentations, COLA tends to perform poorly on speaker recognition tasks.  


Second, we notice that the pretrained APC model~\cite{chung2020generative} helps improve the recognition performance with fine-tuning. For example, compared with the model trained with 1024 speakers from scratch, fine-tuning the APC model with the same labeled dataset improves EER by 24.34\%.

Third, fine-tuning the supervised model pretrained on the VoxCeleb2 dataset also helps improve the EER performance, in spite of the domain mismatch between VoxCeleb2 (YouTube recordings) and Alexa traffic. We observe 31.38\% relative EER improvement when the model is fine-tuned with 1,024 speakers.

Fourth, the proposed method achieves the largest relative EER improvements on all four fine-tuning datasets compared to COLA, APC, and the supervised model trained on the VoxCeleb2 dataset. The best results are highlighted in bold in Table \ref{table:finetuning}.
This demonstrates the superiority of the proposed method for our speaker recognition scenario, learning to distinguish speakers by selectively learning from the unlabeled human-machine dialogues.

\section{Conclusions}
\label{sec:conclusion}

We present a self-supervised learning method for speaker recognition tasks designed to exploit implicit speaker identity information in unlabeled human-machine dialogues. We propose an effective soft rejection mechanism to deal with dialogues containing multiple speakers. 
Experiments on deidentified smart-speaker production data show that the proposed algorithm is effective at handling unsupervised speaker information, giving performance comparable to supervised models.  
When used for model pretraining before supervised training, our method reduces EER by up to 41\% relative, compared to no pretraining, and is superior to  other self-supervised pretraining methods, as well as to pretraining on a large labeled (but domain-mismatched) dataset.

\bibliographystyle{IEEEbib}
\bibliography{refs}

\end{document}